\documentclass[a4paper, 10pt, conference]{ieeeconf}      

\IEEEoverridecommandlockouts                              

\usepackage{graphics} 
\usepackage{amsmath}
\usepackage{amsfonts}
\usepackage{amssymb}
\usepackage{color}
\usepackage{graphicx}
\usepackage{booktabs}
\usepackage{cancel}

\def\x{{\mathbf x}}
\def\y{{\mathbf y}}
\def\h{{\mathbf h}}
\def\b{{\mathbf b}}
\def\W{{\mathbf W}}
\def\R{{\mathbf R}}

\definecolor{darkgreen}{rgb}{0,0.6,0.2}
\definecolor{mypink}{RGB}{219, 48, 122}
\definecolor{some_color}{RGB}{123, 20, 80}
\definecolor{some_color1}{RGB}{80, 100, 20}

\title{\LARGE \bf
Classification of postoperative surgical site infections from blood measurements with missing data using recurrent neural networks
}

\author{Andreas Storvik Strauman\textsuperscript{*,1}, Filippo Maria Bianchi\textsuperscript{1}, Karl \O{}yvind Mikalsen\textsuperscript{1}, Michael Kampffmeyer\textsuperscript{1},\\ Cristina Soguero-Ruiz\textsuperscript{1,2} and Robert Jenssen\textsuperscript{1}
\thanks{$^{1}$UiT Machine Learning Group, UiT The Arctic University of Norway, Tromsø, Norway}%
\thanks{$^{2}$Dept. of Signal Theory and Comm., Telematics and Computing, Universidad Rey Juan Carlos, Fuenlabrada, Spain}%
\thanks{\textsuperscript{*}Corresponding author: \texttt{andreas.s.strauman@uit.no}}
\thanks{This work was partially funded by the Norwegian Research Council FRIPRO grant no. 239844 \emph{Next Generation Learning Machines} and IKTPLUSS grant no. 270738 \emph{Deep Learning for Health}. CSR is  supported by projects TEC2016-75361-R from Spanish Government and DTS17/00158 from Institute of Health Carlos III (Spain).}
}

\begin{document}

\maketitle
\thispagestyle{empty}
\pagestyle{empty}

\begin{abstract}
Clinical measurements that can be represented as time series constitute an important fraction of the electronic health records and are often both uncertain and incomplete.
Recurrent neural networks are a special class of neural networks that are particularly suitable to process time series data but, in their original formulation, cannot explicitly deal with missing data.
In this paper, we explore imputation strategies for handling missing values in classifiers based on recurrent neural network (RNN) and apply a recently proposed recurrent architecture, the Gated Recurrent Unit with Decay, specifically designed to handle missing data. 
We focus on the problem of detecting surgical site infection in patients by analyzing time series of their blood sample measurements and
we compare the results obtained with different RNN-based classifiers.
\end{abstract}

\section{INTRODUCTION}

\emph{Surgical Site Infection} (SSI) is one of the most common types of nosocomial infection~\cite{lewis2013} and represents up to 30\% of hospital-acquired infections~\cite{magill2012prevalence}.  
Studies have shown that being infected with SSI both increases the risk of re-admissions~\cite{shah2017evaluation}, and prolongs the postoperative stay for up to two weeks and thereby also the cost per patient~\cite{whitehouse2002}. 
Hence, being able to detect infections is of utmost importance both for the patients and for the healthcare system.



\emph{Blood sample} measurements represent a fundamental source of information for predicting the risk of getting SSI in a given patient. In some studies blood tests have been jointly analyzed together with other electronic health record data for detecting the presence of SSI~\cite{soguero2016support,hu2017strategies}. 
However, due to the fact that blood samples are recorded frequently with low burden for the patients and describe the health status of a patient with certainty, other studies have successfully focused on the analysis of blood tests alone for predicting SSI~\cite{soguero2015data,angiolini2016}.

Blood samples are usually collected for each patient in given periods both before and after the surgery.  
The data are series of several indicators, measured on a patient over time and, due to the presence of important relationships in time among the measurements, data can be naturally represented as \emph{multivariate time series}.
An effective machine learning framework used to model and analyze multivariate time series (MTS) is the \emph{Recurrent Neural Network} (RNN).

RNNs are a special class of Neural Networks characterized by internal self-connections, which are capable of modeling sequential data of variable lengths~\cite{DBLP:journals/corr/BianchiMKRJ17}.
Thanks to their recurrent nature, an RNN captures temporal dependencies in the MTS to perform prediction or classification.
At each time step, the RNN output depends on past inputs and previously computed states. 
This allows the network to develop a memory of previous events, which is implicitly encoded in its internal state. 
Thanks to these properties, RNNs have proven powerful in a number of different healthcare applications~\cite{pmlr-v56-Choi16, GULER2005506}.

Although a vanilla RNN, or Elman RNN (ERNN), can in principle learn how to model very complex relationships, an optimal training is often difficult to achieve and the network often performs poorly on unseen data and fails to capture long-term dependencies.
A more sophisticated architecture, called Gated Recurrent Unit~\cite{cho2014learning} (GRU), implements recurrent units that adaptively captures dependencies at different time scales and it demonstrated to outperform other architectures on several tasks~\cite{DBLP:journals/corr/ChungGCB14}.

Most clinical data, including blood sample measurements, are corrupted by the presence of missing values.
Indeed, for each patient some measurements may not be registered and, at some time, data might be not collected at all. 
Most machine learning models, including RNNs, are not designed to deal with missing data and their presence often complicates the training and deteriorate performances~\cite{García-Laencina2010}.
A commonly used approach is to replace missing data with imputation methods~\cite{AZPH:AZPH464}, trying to introduce as less bias as possible.


The \emph{Gated Recurrent Unit with Decay} (GRUD)~\cite{DBLP:journals/corr/ChePCSL16} is a recently proposed RNN, specifically designed to handle MTS with missing data and to leverage on the missing patterns to achieve better prediction results.
GRUD takes as input two representations of missing patterns: a \textit{masking} that informs the model which inputs are observed or missing and \textit{time intervals} that encapsulate the input observation patterns.

In this work, we study the problem of identifying surgical site infection by only relying on blood measurements that contain many missing data. 
We evaluate the classification performance of three types of RNN-based classifiers, to discriminate between MTS relative to infected and non-infected patients.
Specifically, we compare ERNN and GRU, where missing data are imputed, and GRUD, which handles missing data without having to resort to imputation.

\section{METHODS}

Let us consider a dataset of $N$ multivariate time series with $V$ variables of same length $T$.
Since a time series $\mathbf{X}_n \in \mathbb{R}^{T \times V}$ may contain missing entries, according to the procedure in~\cite{DBLP:journals/corr/ChePCSL16} we associate to $\mathbf{X}_n$ a binary \textit{mask} $\mathbf{M}_n \in \mathbb{R}^{T \times V}$, whose element $m_t^v = 0$ if $x_t^v$ is missing, otherwise $m_t^v = 1$.

\subsection{Approaches for handling missing data}
\label{sec:inputation}
To replace missing values from the input data, we consider three baseline imputation techniques \cite{AZPH:AZPH464}.
\begin{itemize}
    \item \textit{Zero imputation}: the missing values in each time series are replaced with $0$. The main drawback of this imputation is the introduction of a strong bias in the data.
    \item \textit{Last value carried forward}: for each variable $v$ in $\mathbf{X}_n$ the missing values are replaced by the last value observed for $v$. The main problem with this method is the assumption that there will be no change from one observed value to the next.
    \item \textit{Mean substitution}: for each variable $v$ in $\mathbf{X}_n$ the missing values are replaced by the mean value of $v$ across all the $N$ time series. Mean values are computed only relative to values that are observed, i.e. that are associated to a ``1'' in each matrix $\mathbf{M}$. As main drawback, this method can lead to under-estimates of the variance.
\end{itemize}

\subsection{Elman RNN}
The state update in a ERNN is governed by the difference equation $\h_t = f \left( \W_h \h_{t-1} + \W_i \x_t +\b_h \right)$, where $\W_h$ and $\W_i$ are the recurrent and inputs weights respectively, $\b_h$ is a bias vector, and $f()$ is the activation function usually implemented by a \textit{tanh}.
The network output is computed as $\hat \y = g\left(\W_o \h_T + \b_o \right)$, where $\h_T$ is the last hidden state of the RNN produced once the whole MTS is processed, $\W_o$ and $\b_o$ are the output weights and bias respectively, and $g()$ is a softmax function.
The parameters $\W_i$, $\W_h$, $\W_o$, and $\b_o$ are trained with gradient descent so that the $\hat \y$ matches a desired output $\y$.

\subsection{Gated Recurrent Unit}
\begin{figure}[th!]
    \centering
    \includegraphics[keepaspectratio,width=0.6\columnwidth]{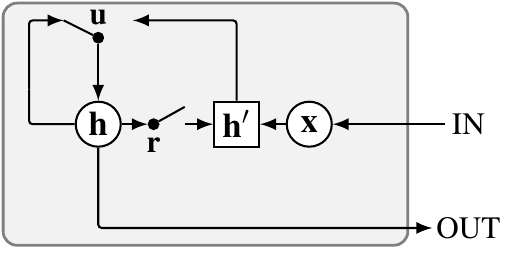}
    \caption{Schema of the GRU cell.}
    \label{fig:gru}
\end{figure}

The Gated Recurrent Unit (GRU)~\cite{cho2014learning} is a gated architecture that can store information for longer periods of time, with respect to ERNNs. 
While the ERNN neuron implements a single squashing nonlinearity, GRU has a more elaborated processing unit called \textit{cell}, which is composed of different nonlinear components interacting with each other in a particular way.
The internal state of a cell is modified by the network only through linear interactions.
This permits information to backpropagate smoothly across time, with a consequent enhancement of the memory capacity of the cell.
A schema of the GRU cell is depicted in Fig. \ref{fig:gru}.

A GRU protects and controls the information in the cell through two gates.
The \textit{update gate}, controls how much the current content of the cell should be updated with the new candidate state.
The \textit{reset gate} if closed (value near to 0) can effectively reset the memory of the cell and make the unit act as if the next processed input was the first in the sequence. 
The activation of each gate depends on the current external input, the previous state of the GRU cells and their output. 
The state equations of the GRU are the following:
\begin{equation*}
\begin{aligned}
\text{reset gate}:	\; & \mathbf{r}_t = \sigma \left( \W_{r} \h_{t-1} + \R_{r} \x_t +\b_r \right) \\
\text{current state}:	\; & \mathbf{h'}_t = g\left(\W(\h_{t-1} \odot \mathbf{r}_t) + \R \x_t + \b \right) \\
\text{update gate}:	\; & \mathbf{u}_t = \sigma \left(\W_u \h_{t-1} + \R_u \x_t + \b_u \right) \\
\text{new state}:	\; & \h_t = (1-\mathbf{u}_t) \odot \h_{t-1} + \mathbf{u}_t \odot \mathbf{h'}_t
\end{aligned}
\end{equation*}
Here, $g(\cdot)$ and $\sigma(\cdot)$ are a non-linear functions usually implemented as hyperbolic tangent and logistic function, respectively. 
The parameters are the rectangular matrices $\W_r$, $\W$, $\W_u$, the square matrices $\R_r$, $\R$, $\R_u$, and the bias vectors $\b_r$, $\b$, $\b_u$. 
To control the behavior of each gate, those parameters are trained with gradient descent to solve a target task.


\subsection{Gated Recurrent Unit with Decay}

In the GRUD cell the standard GRU architecture is modified to implement a decay mechanism for the input variables and the hidden states, according to the missing values in input.
Such decays capture two different properties that characterize health care data.
First, the values of the missing variable tend to be close to some default value if its last observation occurs far in time~\cite{vodovotz2013systems}.
Second, the influence of the last seen input variables diminish over time when the next values are missing~\cite{ZHOU2007183}.

Beside the mask $\mathbf{M}_n$, to track missing values in each MTS $\mathbf{X}_n$, GRUD maintains the last time interval when each variable $v$ was observed in a matrix $\boldsymbol{\Delta} \in \mathbb{R}^{T \times V}$.
Specifically, an element $\delta_t^v$ of $\boldsymbol{\Delta}$ is defined as 
\[
\delta_t^v = 
\begin{cases}
s_t - s_{t-1} + \delta_{t-1}^v, & t > 1, m_{t-1}^d = 0 \\
s_t - s_{t-1}, & t > 1, m_{t-1}^d = 1 \\
0, & t=1
\end{cases}
\]
where $s_t$ are the time stamps relative to each measurement.
A vector of decay rates $\gamma$ is defined as
\begin{equation}
    \label{eq:decay}
    \gamma_t = \mathrm{exp}\left\{ -\mathrm{max} \left( \mathbf{0}, \mathbf{W}_\gamma \boldsymbol{\delta}_t + \mathbf{b}_\gamma  \right) \right\},
\end{equation}
where $\mathbf{W}_\gamma$ and $\mathbf{b}_\gamma$ are trained on data along with the other parameters.
GRUD employs two different decays. First, $\gamma_x$, decays the input over time toward its empirical mean
\begin{equation}
    \label{eq:input_decay}
    x_t^v \leftarrow m_t^v x_t^v + (1 - m_t^v)\gamma_{xt}^v x_{t'}^v + (1-m_t^v)(1-\gamma_{xt}^v)\tilde{x}^d \; ,
\end{equation}
where $x_{t'}^v$ is the last value observed for variable $v$ and $\tilde{x}^d$ is its empirical mean.
Secondly, $\gamma_h$ decays the extracted features before computing the next hidden state
\begin{equation}
    \label{eq:state_decay}
    \mathbf{h}_{t-1} \leftarrow \gamma_{ht} \odot \mathbf{h}_{t-1} \; .
\end{equation}

\begin{figure}[th!]
    \centering
    \includegraphics[keepaspectratio,width=0.6\columnwidth]{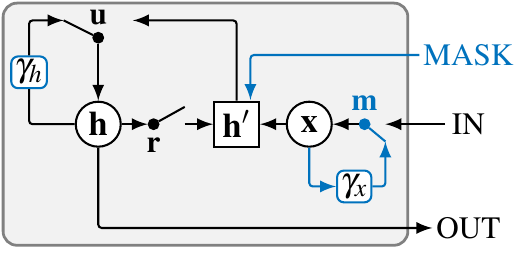}
    \caption{Schema of the GRUD cell. Modification with respect to the original GRU architecture are highlighted in blue.}
    \label{fig:grud}
\end{figure}

The state update equations for GRUD are
\begin{equation*}
\begin{aligned}
& \mathbf{r}_t = \sigma \left( \W_{r} \h_{t-1} + \R_{r} \x_t + \mathbf{V}_r\mathbf{m}_t + \b_r \right) \\
& \mathbf{h'}_t = g\left(\W(\h_{t-1} \odot \mathbf{r}_t) + \R \x_t + \mathbf{V}\mathbf{m}_t + \b \right) \\
& \mathbf{u}_t = \sigma \left(\W_u \h_{t-1} + \R_u \x_t + \mathbf{V}_u\mathbf{m}_t + \b_u \right) \\
& \h_t = (1-\mathbf{u}_t) \odot \h_{t-1} + \mathbf{u}_t \odot \mathbf{h'}_t
\end{aligned}
\end{equation*}
where $\x_t$ and $\h_{t-1}$ are respectively updated according to Eq.~\ref{eq:input_decay} and Eq.~\ref{eq:state_decay}, while $\mathbf{V}_r$, $\mathbf{V}$ and $\mathbf{V}_u$ are additional trainable weights for the masking values $\mathbf{m}_t$ in $\mathbf{M}$.  
A schema of the GRUD architecture is depicted in Fig. \ref{fig:grud}.

Contrarily to the GRU and ERNN, in GRUD it is not necessary to apply an imputation on the input data and the model can be trained end-to-end in presence of missing values.

\subsection{Loss function}

In all three RNNs, the weights are trained using the same loss function, implementing binary cross-entropy combined with a $\mathrm{L}_2$ regularization term. 
Due to the class imbalance in the dataset, we implemented a weighting scheme to penalize mistakes on the minority class by an amount proportional to how under-represented it is.
In particular, errors relative to class $i$ are weighted by a term $\alpha_i = 1 - n_i/N$, where $n_i$ is the number of training samples of class $i$ and $N$ the size of the training set.
In this way, classification errors on the class with less elements contribute more than errors on the other class.
The resulting loss function is
\[
L = -\frac{1}{N}\sum \limits_{i=1}^N \alpha_i \bigg[y_i  \log \hat y_i + (1 - y_i)  \log (1 - \hat y_i)\bigg] + \lambda \| \mathcal{W} \|_2,
\]
where $y_i$ and $\hat y_i$ are the true and predicted class respectively, $\|\mathcal{W}\|_2$ is the $\mathrm{L}_2$ norm of all network weights (biases excluded), and $\lambda$ weights the regularization strength.

\section{EXPERIMENTS}

The purpose of the current study is to discriminate with RNN-based classifiers between MTS of blood samples relative to patients with and without surgical site infection.
The blood samples are continuous variables over time and represented as MTS. 
In our analysis, we discretized time and let each time interval be one day. 
Ten different blood tests were extracted over 20 days after surgery, namely, alanine aminotransferase, albumin, alkaline phosphatase, creatinine, CRP, hemoglobine, leukocytes, potassium, sodium and thrombocytes.

The dataset consist of patients that underwent a gastrointestinal surgical procedure  at UNN in the years 2004 - 2012. To extract the cohort for this study, we considered both the International Classification of Diseases and NOMESCO Classification of Surgical Procedures codes related to severe postoperative complications. A patient that did not have any of these codes was considered as a control, otherwise, as a case.  
We removed patients with less than two measurements during the postoperative window from the cohort. We ended up with a total of 232 infected patients (cases) and 651 not infected (control). 

20 \% of the dataset was used for validation. The remaining part was randomly split into a training (60 \%) set and a test set. 
This procedure was repeated 10 times, using each time a new random initialization of the parameters in the RNNs. 
To measure performance we used F1-score and area under the ROC curve (AUC), which are more suitable performance measures in presence of imbalanced data~\cite{LOPEZ2013113}.

\subsection{Network configuration}

In the experiments, we used identical network architectures and only switched the internal processing units to be ERNN, GRU or GRUD.
More specifically, we used a network with a single layer and $22$ hidden units.
On the output layer we applied dropout with probability $0.2$ and we set the regularization parameter $\lambda = 0.001$.
To train model parameters we used mini batches of size 40 and Adam as optimization algorithm.
Each network is trained for $10,000$ epochs, with data shuffled each time.
The models used for testing are the ones yielding the best F1 score on the validation set.

\subsection{Results}

In Table~\ref{tab:res} we report the mean classification results and standard errors obtained by a RNN classifier configured either with ERNN, GRU or GRUD on the validation set during training and on the final classification of the test set once the training is over.
When using ERNN and GRU, missing values in the inputs are filled using mean substitution (-m), zero imputation (-z) or last value carried forward (-l).

\bgroup
\def\arraystretch{1.2} 
\setlength\tabcolsep{.4em} 
\begin{table}[!ht]
\footnotesize
\centering
\caption{F1 score and area under the ROC curve (AUC) achieved on validation (val) and test by ERNN, GRU and GRUD. 
In ERNN and GRU we used three different imputations: mean imputation (-m), zero imputation (-z) and last value carried forward (-l). Best average results are in bold.}
\label{tab:res}
\begin{tabular}{l|llll}
\cmidrule[1.5pt]{1-5}
\textbf{Model} & \textbf{AUC} (val) & \textbf{F1} (val) & \textbf{AUC} (test) & \textbf{F1} (test) \\
\cmidrule[.5pt]{1-5}
ERNN-m   & $0.76 \pm 0.02$ & $0.31 \pm 0.13$ & $0.76  \pm 0.02$ & $0.33 \pm 0.16$ \\
ERNN-l   & $0.83 \pm 0.02$ & $0.54 \pm 0.05$ & $0.84 \pm 0.02$ & $0.57 \pm 0.06$  \\
ERNN-z   & $0.86 \pm 0.01$ & $0.60 \pm 0.06$ & $0.86 \pm 0.01$ & $0.63 \pm 0.04$  \\
GRU-m    & $0.90 \pm 0.01$ & $0.74 \pm 0.02$ & $0.90 \pm 0.01$ & $0.70 \pm 0.06$ \\
GRU-l    & $0.90 \pm 0.02$ & $0.73 \pm 0.04$ & $0.90 \pm 0.02$ & $0.68 \pm 0.06$ \\
GRU-z    & $0.90 \pm 0.02$ & $0.73 \pm 0.04$ & $0.89 \pm 0.02$ & $0.69 \pm 0.03$ \\
GRUD     & $\mathbf{0.91} \pm 0.02$ & $\mathbf{0.75} \pm 0.05$ & $\mathbf{0.91} \pm 0.02$  & $\mathbf{0.70} \pm 0.05$ \\
\cmidrule[1.5pt]{1-5}
\end{tabular}
\end{table}
\egroup

As we can see, the best classification results in terms of F1-score and AUC are achieved by GRUD and GRU configured with mean imputation, which achieve similar performance in both validation and testing.
Interestingly, GRUD can handle missing values as well as the standard GRU cell and provides the advantage of an end-to-end training, without requiring imputation procedures to be applied in advance.
On the other hand, the ERNN configurations perform worse than the other architectures. 
This is somehow expected, since the presence of the gating mechanisms in GRU and GRUD provide more flexibility and computational capability.

\begin{figure}[th!]
    \centering
    \includegraphics[keepaspectratio,width=\columnwidth]{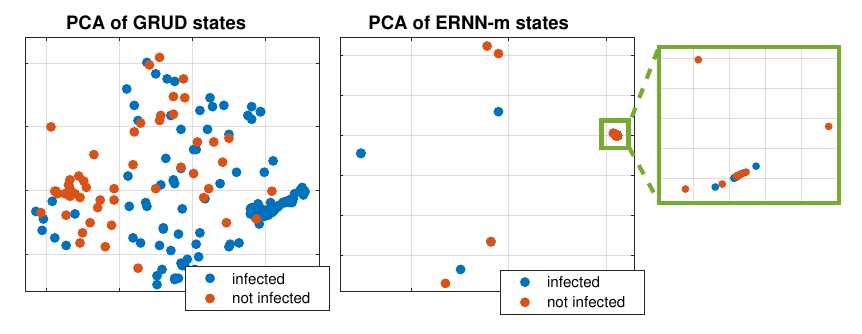}
    \caption{First two principal components of the last states in GRUD and ERNN, when the MTS in the test set are processed. ERNN states relative to different classes are highly overlapped, as shown in the green box zoom.}
    \label{fig:pca}
\end{figure}

To analyze the quality of the representations learned by GRUD compared to ERNN, we performed principal component analysis (PCA) on the final hidden states of the networks.
The classification outcome heavily depends on those states since they are the input to the last softmax layer, which produces class assignment.
Each last state is therefore the high-level static representation of the sequential input learned by the network.
In Fig.~\ref{fig:pca} the representations relative to the MTS in the test set are mapped to two dimensions. 
As we can see, the GRUD separates the two classes well, while in the case of ERNN the test elements are highly overlapped. 

We observe that, in contrast to GRU, the performance of ERNN is heavily affected by the choice of imputation method.
Indeed, each technique introduces a different kind of bias in the data and the optimal choice depends on type of task at hand.
Using the wrong imputation may complicate the training.
While this can represent an issue in the weaker ERNN, the higher computational capability of GRU permits to handle well the presence of stronger biases.

\section{CONCLUSIONS}

In this work we focused on the classification of blood sample data relative to patients with surgical site infections.
Data are represented by multivariate time series and are characterized by a large amount of missing values.
To classify the data, we used three different RNNs configured either with ERNN, GRU or GRUD.
While GRUD can process MTS with missing values, ERNN and GRU require imputation to replace missing values.
In the experiments, we observed that GRUD and GRU with imputation achieves better performance than ERNN in classifying the MTS.
We also noticed that different imputations yield a substantial variation in ERNN classification results, while the performance in GRU are more stable.
Since selecting the best imputation method is often difficult and requires expertise on the data domain, a critical sensitivity in this configuration may represent an issue.
Therefore, the stability provided by GRU and the GRUD, which does not require using imputation at all, is an important advantage in many practical applications.


\bibliographystyle{IEEEtran}
\bibliography{IEEEabrv,biblio}

\begin{thebibliography}{10}
\providecommand{\url}[1]{#1}
\csname url@samestyle\endcsname
\providecommand{\newblock}{\relax}
\providecommand{\bibinfo}[2]{#2}
\providecommand{\BIBentrySTDinterwordspacing}{\spaceskip=0pt\relax}
\providecommand{\BIBentryALTinterwordstretchfactor}{4}
\providecommand{\BIBentryALTinterwordspacing}{\spaceskip=\fontdimen2\font plus
\BIBentryALTinterwordstretchfactor\fontdimen3\font minus
  \fontdimen4\font\relax}
\providecommand{\BIBforeignlanguage}[2]{{%
\expandafter\ifx\csname l@#1\endcsname\relax
\typeout{** WARNING: IEEEtran.bst: No hyphenation pattern has been}%
\typeout{** loaded for the language `#1'. Using the pattern for}%
\typeout{** the default language instead.}%
\else
\language=\csname l@#1\endcsname
\fi
#2}}
\providecommand{\BIBdecl}{\relax}
\BIBdecl

\bibitem{lewis2013}
S.~S. Lewis, R.~W. Moehring, L.~F. Chen \emph{et~al.}, ``Assessing the relative
  burden of hospital-acquired infections in a network of community hospitals,''
  \emph{Infection Control \& Hospital Epidemiology}, 2013.

\bibitem{magill2012prevalence}
S.~S. Magill, W.~Hellinger, J.~Cohen \emph{et~al.}, ``Prevalence of
  healthcare-associated infections in acute care hospitals in {J}acksonville,
  {F}lorida,'' \emph{Infection Control}, 2012.

\bibitem{shah2017evaluation}
R.~Shah, E.~Pavey, M.~Ju \emph{et~al.}, ``Evaluation of readmissions due to
  surgical site infections: A potential target for quality improvement,''
  \emph{The American Journal of Surgery}, 2017.

\bibitem{whitehouse2002}
J.~D. Whitehouse, N.~D. Friedman, K.~B. Kirkland, W.~J. Richardson, and D.~J.
  Sexton, ``The impact of surgical-site infections following orthopedic surgery
  at a community hospital and a university hospital adverse quality of life,
  excess length of stay, and extra cost,'' \emph{Infection Control \& Hospital
  Epidemiology}, 2002.

\bibitem{soguero2016support}
C.~Soguero-Ruiz, K.~Hindberg, R.~Jenssen \emph{et~al.}, ``Support vector
  feature selection for early detection of anastomosis leakage from
  bag-of-words in electronic health records,'' \emph{IEEE journal of biomedical
  and health informatics}, 2016.

\bibitem{hu2017strategies}
Z.~Hu, G.~B. Melton, E.~G. Arsoniadis, Y.~Wang, M.~R. Kwaan, and G.~J. Simon,
  ``Strategies for handling missing clinical data for automated surgical site
  infection detection from the electronic health record,'' \emph{Journal of
  Biomedical Informatics}, 2017.

\bibitem{soguero2015data}
C.~Soguero-Ruiz, W.~M. Fei, R.~Jenssen \emph{et~al.}, ``Data-driven temporal
  prediction of surgical site infection,'' in \emph{AMIA Annual Symposium
  Proceedings}, vol. 2015.\hskip 1em plus 0.5em minus 0.4em\relax American
  Medical Informatics Association, 2015, p. 1164.

\bibitem{angiolini2016}
M.~R. Angiolini, F.~Gavazzi, C.~Ridolfi, M.~Moro, P.~Morelli, M.~Montorsi, and
  A.~Zerbi, ``Role of c-reactive protein assessment as early predictor of
  surgical site infections development after pancreaticoduodenectomy,''
  \emph{Digestive surgery}, 2016.

\bibitem{DBLP:journals/corr/BianchiMKRJ17}
F.~M. Bianchi, E.~Maiorino, M.~C. Kampffmeyer, A.~Rizzi, and R.~Jenssen, ``An
  overview and comparative analysis of recurrent neural networks for short term
  load forecasting,'' \emph{CoRR}, vol. abs/1705.04378, 2017.

\bibitem{pmlr-v56-Choi16}
E.~Choi, M.~T. Bahadori, A.~Schuetz, W.~F. Stewart, and J.~Sun, ``Doctor {AI}:
  Predicting clinical events via recurrent neural networks,'' in \emph{Proc.
  1st Machine Learning for Healthcare Conf.}\hskip 1em plus 0.5em minus
  0.4em\relax PMLR, 2016.

\bibitem{GULER2005506}
N.~F. Güler, E.~D. Übeyli, and İnan Güler, ``Recurrent neural networks
  employing lyapunov exponents for eeg signals classification,'' \emph{Expert
  Systems with Applications}, 2005.

\bibitem{cho2014learning}
K.~Cho, B.~Van~Merri{\"e}nboer, C.~Gulcehre, D.~Bahdanau, F.~Bougares,
  H.~Schwenk, and Y.~Bengio, ``Learning phrase representations using rnn
  encoder-decoder for statistical machine translation,'' \emph{arXiv preprint
  arXiv:1406.1078}, 2014.

\bibitem{DBLP:journals/corr/ChungGCB14}
J.~Chung, {\c{C}}.~G{\"{u}}l{\c{c}}ehre, K.~Cho, and Y.~Bengio, ``Empirical
  evaluation of gated recurrent neural networks on sequence modeling,''
  \emph{CoRR}, 2014.

\bibitem{García-Laencina2010}
P.~J. Garc{\'i}a-Laencina, J.-L. Sancho-G{\'o}mez, and A.~R. Figueiras-Vidal,
  ``Pattern classification with missing data: a review,'' \emph{Neural
  Computing and Applications}, 2010.

\bibitem{AZPH:AZPH464}
D.~A. Bennett, ``How can i deal with missing data in my study?''
  \emph{Australian and New Zealand Journal of Public Health}, 2001.

\bibitem{DBLP:journals/corr/ChePCSL16}
Z.~Che, S.~Purushotham, K.~Cho, D.~Sontag, and Y.~Liu, ``Recurrent neural
  networks for multivariate time series with missing values,'' \emph{CoRR},
  vol. abs/1606.01865, 2016.

\bibitem{vodovotz2013systems}
Y.~Vodovotz, G.~An, and I.~P. Androulakis, ``A systems engineering perspective
  on homeostasis and disease,'' \emph{Frontiers in bioengineering and
  biotechnology}, 2013.

\bibitem{ZHOU2007183}
L.~Zhou and G.~Hripcsak, ``Temporal reasoning with medical data $-$ a review
  with emphasis on medical natural language processing,'' \emph{Journal of
  Biomedical Informatics}, 2007.

\bibitem{LOPEZ2013113}
V.~Lopez, A.~Fernández, S.~García \emph{et~al.}, ``An insight into
  classification with imbalanced data: Empirical results and current trends on
  using data intrinsic characteristics,'' \emph{Information Sciences}, 2013.

\end{thebibliography}

\end{document}